\documentclass[10pt,twocolumn,letterpaper]{article}

\usepackage{iccv}      

\definecolor{iccvblue}{rgb}{0.21,0.49,0.74}
\usepackage{caption}
\usepackage[pagebackref,breaklinks,colorlinks,allcolors=iccvblue]{hyperref}
\usepackage{float}
\usepackage{algorithm}
\usepackage{algorithmic}
\usepackage{bbding}
\usepackage{array}
\usepackage{makecell} 
\usepackage{graphicx}
\usepackage{subcaption}

\usepackage{marvosym}

\usepackage[margin=1in]{geometry}
\usepackage{xcolor} 
\newcolumntype{C}[1]{>{\centering\arraybackslash}m{#1}}

\def\paperID{*****} 
\def\confName{ICCV}
\def\confYear{2025}

\title{Bind-Your-Avatar: Multi-Talking-Character Video Generation with Dynamic 3D-mask-based Embedding Router}

\author{
Yubo Huang$^{1}$ \quad
Weiqiang Wang$^{2}$ \quad
Sirui Zhao$^{1\,\textsuperscript{\Letter}}$ \quad
Tong Xu$^{1}$ \\
Lin Liu$^{1\,\textsuperscript{\Letter}}$ \quad
Enhong Chen$^{1\,\textsuperscript{\Letter}}$ \\
$^{1}$University of Science and Technology of China \\
$^{2}$Monash University \\
\small{\textsuperscript{\Letter} Corresponding Authors. siruit@ustc.edu.cn, {ll0825}@mail.ustc.edu.cn, cheneh@ustc.edu.cn}
}
\begin{document}

\twocolumn[{%
\renewcommand\twocolumn[1][]{#1}%
\maketitle

\begin{center}
    \centering
    \includegraphics[width=1.1\textwidth]{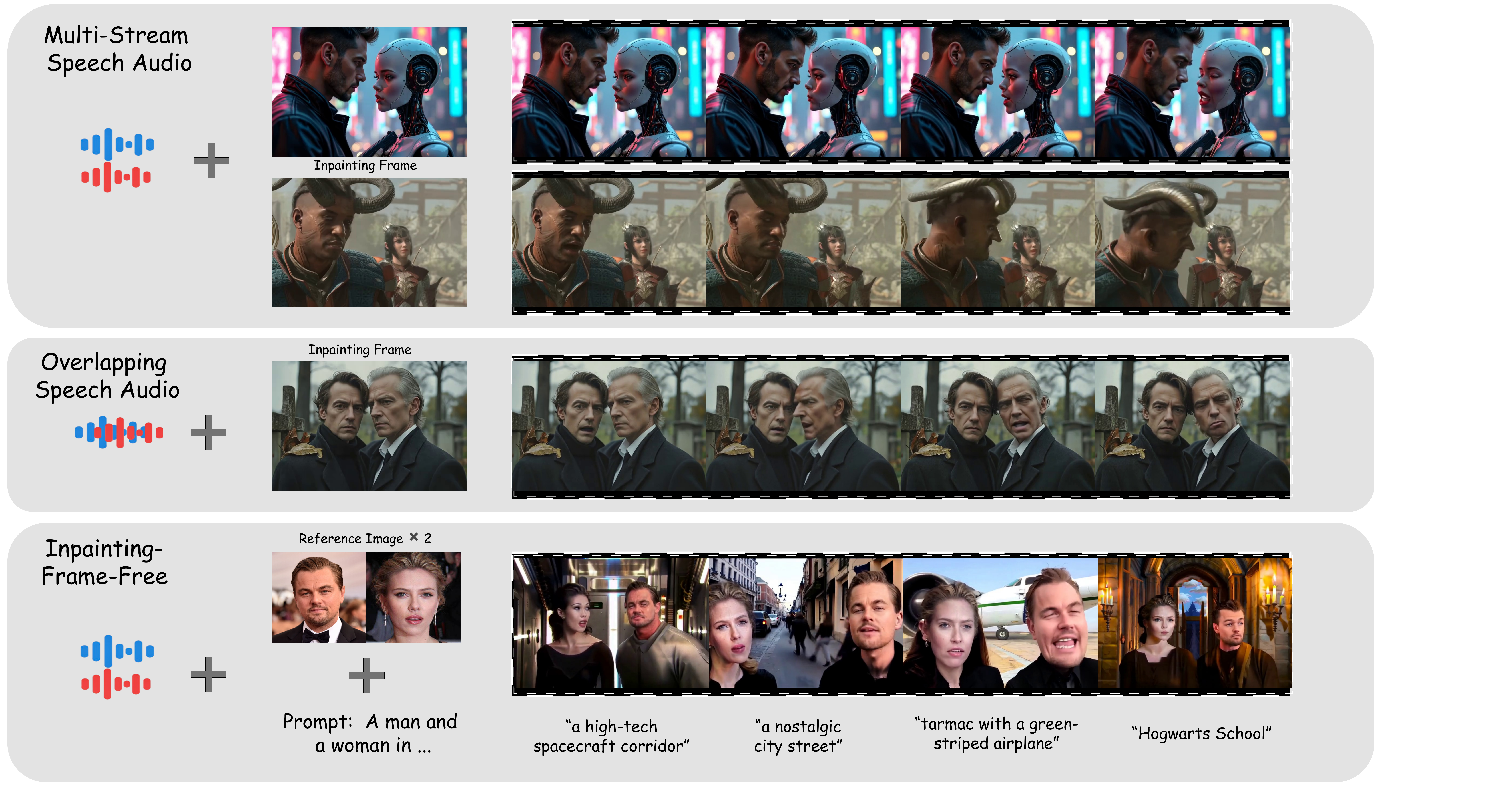}
    \captionof{figure}{\textbf{Bind-Your-Avatar} is a novel framework that generates videos from multi-stream speech audio with precise lip-sync for each speaker. It robustly handles overlapping speech and supports prompt-driven, inpainting-free multi-talking-characrer video generation. \href{https://yubo-shankui.github.io/bind-your-avatar/}{Page: https://yubo-shankui.github.io/bind-your-avatar/}}
\end{center}%
}]

\begin{abstract}
Recent years have witnessed remarkable advances in audio-driven talking head generation. However, existing approaches predominantly focus on single-character scenarios. While some methods can create separate conversation videos between two individuals, the
critical challenge of generating unified conversation videos
with multiple physically co-present characters sharing the
same spatial environment remains largely unaddressed.
This setting presents two key challenges: audio-to-character correspondence control and the lack of suitable datasets featuring multi-character talking videos within the same scene. To address these challenges, we introduce Bind-Your-Avatar, an MM-DiT-based model specifically designed for multi-talking-character video generation in the same scene.
Specifically, we propose (1) A novel framework incorporating a fine-grained Embedding Router that \textbf{binds} \textbf{``who''} and \textbf{``speak what''} together to address the audio-to-character correspondence control. (2) Two methods for implementing a 3D-mask embedding router that enables frame-wise, fine-grained control of individual characters, with distinct loss functions based on observed geometric priors and a mask refinement strategy to enhance the accuracy and temporal smoothness of the predicted masks. (3) The first dataset, to the best of our knowledge, specifically constructed for multi-talking-character video generation, and accompanied by an open-source data processing pipeline, and (4) A benchmark for the dual-talking-characters video generation, with extensive experiments demonstrating superior performance over multiple state-of-the-art methods. 
\end{abstract}    

\section{Introduction} 
Generating animatable avatars from static portraits through diverse control signals (including audio, facial keypoints, text prompts, and dense motion flow) represents a fundamental challenge in computer vision and graphics. This technology has broad applications across gaming, filmmaking, and virtual reality.
Among animation techniques driven by these control signals, audio-driven animation (commonly termed talking head generation) has emerged as particularly valuable and widely studied.
It focuses on synthesizing realistic digital characters whose facial expressions, lip movements, and head poses are precisely synchronized with input speech audio. 
Recent sophisticated methods have made significant strides in high-fidelity talking portrait generation, with recent advances largely powered by diffusion models 
\cite{chenEchoMimicLifelikeAudioDriven2024,cuiHallo2LongDurationHighResolution2024,cuiHallo3HighlyDynamic2025,jiSonicShiftingFocus2024,linOmniHuman1RethinkingScalingUp2025,tianEMO2EndEffectorGuided2025,mengEchoMimicV2StrikingSimplified2024,yangMegActor$S$UnlockingFlexible2024,weiMoChaMovieGradeTalking2025,yiMagicInfiniteGeneratingInfinite2025,zhangLetsTalkLatentDiffusion2024,zhenTellerRealTimeStreaming2025,wangFantasyTalkingRealisticTalking2025,wangJoyGenAudioDriven3D2025}
, yet remain constrained primarily to single-character scenarios. Although some approaches generate separate multi-talking-character videos featuring isolated individuals in distinct scenes 
\cite{qiChatAnyoneStylizedRealtime2025a,zhuINFPAudioDrivenInteractive2024,parkLetsGoReal2024}, the critical challenge of generating unified multi-talking-character videos with multiple physically co-present characters sharing the same spatial environment remains largely unaddressed.
\begin{table}[t]
    \centering
    \caption{Comparison of Supported Input Modalities in State-of-the-Art Character Animation Methods.}
    \label{tab:comparison-of-supported}
    \fontsize{6}{6.5}\selectfont
    \setlength{\tabcolsep}{3pt}
    \renewcommand{\arraystretch}{1.4}
    \begin{tabular}{p{1.7cm}C{1.2cm}C{1.3cm}C{1.3cm}C{1.3cm}}
        \hline\noalign{\hrule height 0.6pt}
        \textbf{Method} 
        & \makecell{\textbf{Speech} \\[-2pt] } 
        & \makecell{\textbf{Multi-stream} \\[1pt] \textbf{Speech }} 
        & \makecell{\textbf{Overlapping-} \\[-0pt] \textbf{Speech }} 
        & \makecell{\textbf{Inpainting} \\[-0pt] \textbf{Frame-Free}} \\
        \hline
        ConsisID \cite{yuan_identity-preserving_2024} & \XSolidBrush & \XSolidBrush & \XSolidBrush & \Checkmark \\
        Ingredients \cite{fei_ingredients_2025} & \XSolidBrush & \XSolidBrush & \XSolidBrush & \Checkmark \\
        Hallo3 \cite{cuiHallo3HighlyDynamic2025} & \Checkmark & \XSolidBrush & \XSolidBrush & \XSolidBrush \\
        Sonic \cite{jiSonicShiftingFocus2024} & \Checkmark & \XSolidBrush & \XSolidBrush & \XSolidBrush \\
        Mocha \cite{weiMoChaMovieGradeTalking2025} & \Checkmark & \XSolidBrush & \XSolidBrush & \Checkmark \\
        MultiTalk \cite{kong2025let} & \Checkmark & \Checkmark & \XSolidBrush & \XSolidBrush \\
        HunyuanAvatar \cite{chenHunyuanVideoAvatarHighFidelityAudioDriven2025} & \Checkmark & \Checkmark & \XSolidBrush & \XSolidBrush \\
        \textbf{Ours} & \Checkmark & \Checkmark & \Checkmark & \Checkmark \\
        \hline\noalign{\hrule height 0.6pt}
    \end{tabular}
\end{table}

In this paper, we advance the study of generating unified multi-talking-character videos featuring two physically interacting characters within the same scene, conditioned on multi-stream speech input, multiple reference character images, and natural language descriptions. We refer to this new setting as the task of Multi-Talking-Character Video Generation.
Unlike traditional talking-head generation, which focuses solely on single-character scenarios, this task involves driving conditions for multiple characters, 
which introduces the critical need to accurately bind each speech audio to its corresponding character.
Additionally, while traditional talking-head generation typically uses a reference character image as the first frame of the generated video\footnote{Hereafter, we term this image an inpainting frame.}, this practice does not apply to the multi-character setting, as we cannot simply concatenate multiple character reference images to create an inpainting frame. Instead, natural-language description is utilized to compose the initial frame. As summarized in Table \ref{tab:comparison-of-supported}, compared with other audio-driven multi-modal methods, our approach uniquely supports more flexible input modalities critical for this novel task.

This task presents two key challenges: (1) the precise control of audio-to-character correspondence, ensuring that each character is animated exclusively by their respective speech audio; and (2) the lack of suitable datasets featuring multi-character talking videos within the same scene, which hinders the advancement of effective models in this setting. To address these challenges, we introduce \textbf{Bind-Your-Avatar}, an MM-DiT-based model specifically designed for multi-talking-character video generation in the same scene. 
\begin{figure*}[t]
    \centering
    \includegraphics[width=1.01\textwidth]{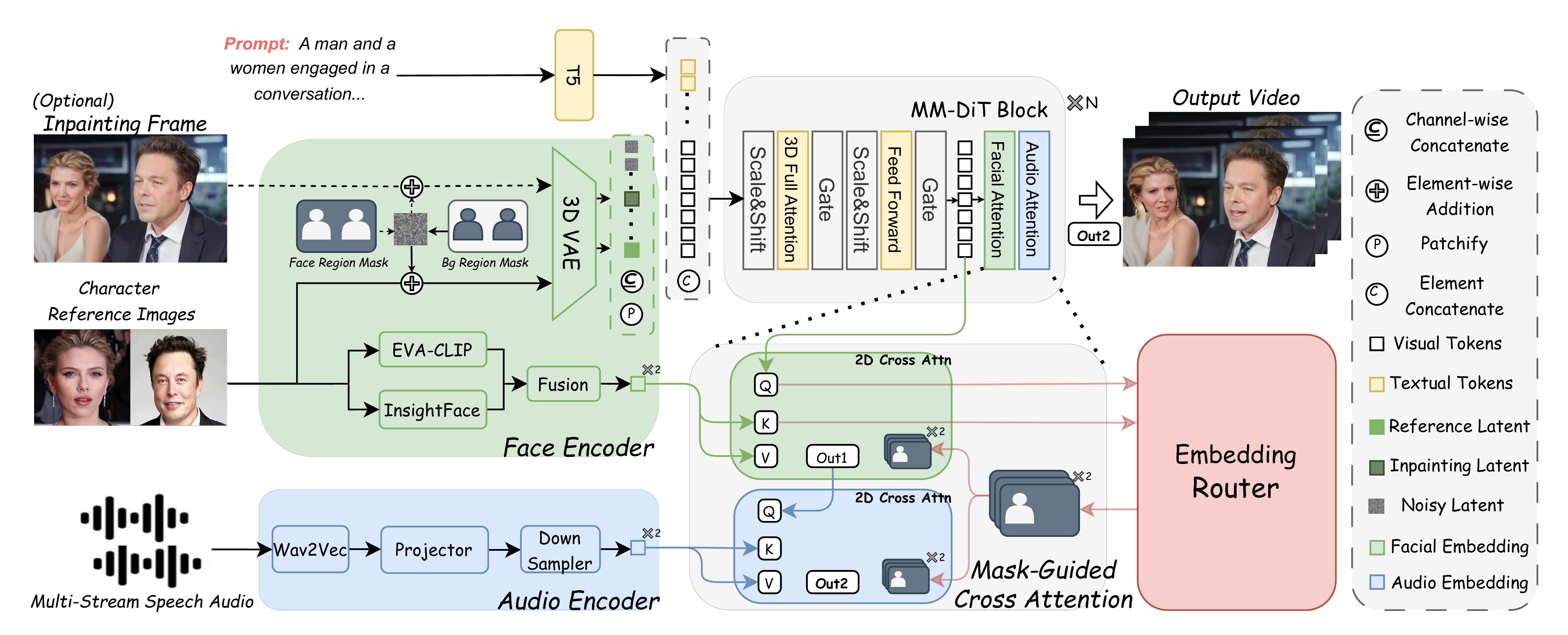} 
    \caption{
    Overview of the proposed framework, which consists of four main components: 
    (1) a \textbf{Multi-Modal Diffusion Transformer \textcolor{gray}{(MM-DiT)}} that generates video sequences conditioned on text, audio, and visual inputs; 
    (2) a \textbf{\textcolor[HTML]{82B366}{Face Encoder}} that extracts facial features;
    (3) an \textbf{\textcolor[HTML]{6C8EBF}{Audio Encoder}} that captures motion-related information; and
    (4) an \textbf{\textcolor[HTML]{B85450}{Embedding Router}} that binds ``who'' and ``speaks what'' together, enabling precise audio-to-character correspondence control.
    Dual mask-guided cross-attention modules in MM-DiT selectively incorporate both motion-related speech information and facial embeddings into visual tokens, using fine-grained masks predicted by the embedding router.
    }
    \label{fig1}
\end{figure*}
For challenge (1), we design a novel framework based on Diffusion Transformer, incorporating a fine-grained Embedding Router that \textbf{binds} speaker identity \textbf{(``who'')} with speech content \textbf{(``what'')} in shared scenes. Three implementations to achieve this router were explored: The Pre-Denoise Router uses face detection from the inpainting frame to guide speech alignment, but fails with dynamic scenes and close interactions. The Post-Denoise Router first generates silent videos to obtain character masks, but introducing costly multi-pass computation. Our optimal Intra-Denoise Router dynamically generates visual masks during the denoising process within a single forward pass, achieving comparable accuracy with minimal overhead.

For challenge (2), we construct a new dataset  designed specifically for multi-talking-character video generation, comprising over 200 hours of high-quality videos curated from diverse sources. 
To the best of our knowledge, there is no existing dataset specifically constructed for this task.
We further develop an automated data processing pipeline to support training for this task, which performs three core operations: (a) separating individual speech audios from multi-character overlapping videos, (b) generating dense segmentation masks for each character, and (c) assigning character identity labels to each audio segment and segmentation mask to specify their corresponding characters.

Our main contributions are as follow: (1) We firstly introduce the multi-talking-character video generation task, analyze its challenges in detail, and propose a novel framework incorporating a Fine-Grained embedding router that binds ``who'' and ``speak what'' together to address the audio-to-character correspondence control in the same scene. 
(2) We propose two methods to implement a 3D-mask embedding router that enables frame-wise fine-grained control. To further ensure the accuracy of the predicted masks, we introduce distinct geometry-based loss functions and a mask refinement strategy.
(3) We release, to the best of our knowledge, the \textbf{first} dataset specifically constructed for multi-talking-character video generation, and will open-source our data cleaning and processing pipeline to facilitate further research in this area.
(4) We establish a benchmark for the dual-talking-character video generation task and demonstrate superior performance compared to multiple state-of-the-art methods.

\section{Related Work}
\subsection{Audio-driven Character Animation}
Pioneering audio-driven character animation utilized multi-stage pipelines with intermediate representations like landmarks, depth maps, and neural fields \cite{zhongIdentityPreservingTalkingFace2023, wilesX2FaceNetworkControlling2018, hongDepthAwareGenerativeAdversarial2022, guoADNeRFAudioDriven2021, zhouyangMakeltTalk2020, prajwalLipSyncExpert2020}. 
These intermediate steps simplified learning by modeling structure explicitly, but introduced error accumulation and reduced visual realism.
Recently, diffusion models make direct synthesis feasible by implicitly learning geometry, identity, and motion, thus removing the need for intermediate steps and enhancing realism. \cite{wangFantasyTalkingRealisticTalking2025, weiMoChaMovieGradeTalking2025, yiMagicInfiniteGeneratingInfinite2025, cuiHallo3HighlyDynamic2025, chenEchoMimicLifelikeAudioDriven2024, yangMegActor$S$UnlockingFlexible2024}. Although achieving impressive results, these methods rely on pre-defined inpainting frames—typically the first frame of the video used to guide generation—limiting scene flexibility. Besides, they focus primarily on single-character scenarios restricting multi-character applications, as shown in Table \ref{tab:comparison-of-supported}. Concurrent works \cite{kong2025let,chenHunyuanVideoAvatarHighFidelityAudioDriven2025} attempt to extend talking head generation to multi-character scenarios, but they still rely on a provided inpainting frame. In addition, their use of static masks limits the modeling of dynamic scenes.
\subsection{Multi-Concept Video Generation}
Recent advances enable multi-concept video generation through specialized architectures. MAGREF \cite{dengMAGREFMaskedGuidance2025} introduces masked guidance for multi-subject synthesis, while DanceTogether \cite{chenDanceTogetherIdentityPreservingMultiPerson2025} preserves identities in interactive scenes. Concat-ID \cite{zhongConcatIDUniversalIdentityPreserving2025} and ConceptMaster \cite{huangConceptMasterMultiConceptVideo2025} achieve universal identity preservation using concatenated embeddings and decoupled concept learning.
Ingredients \cite{fei_ingredients_2025} project facial embeddings and visual tokens into a shared space to compute relevance scores for assigning each token to a character. However, the resulting masks lack precision and temporal smoothness due to the absence of geometric priors that naturally constrain 3D mask shapes.
Moreover, these methods primarily rely on image and text conditioning and lack support for audio-driven multi-character generation. This limitation restricts their applicability in scenarios that require audio-synchronized narrative content creation.
\begin{figure*}[t]
    \centering
    \includegraphics[width=1.01\textwidth]{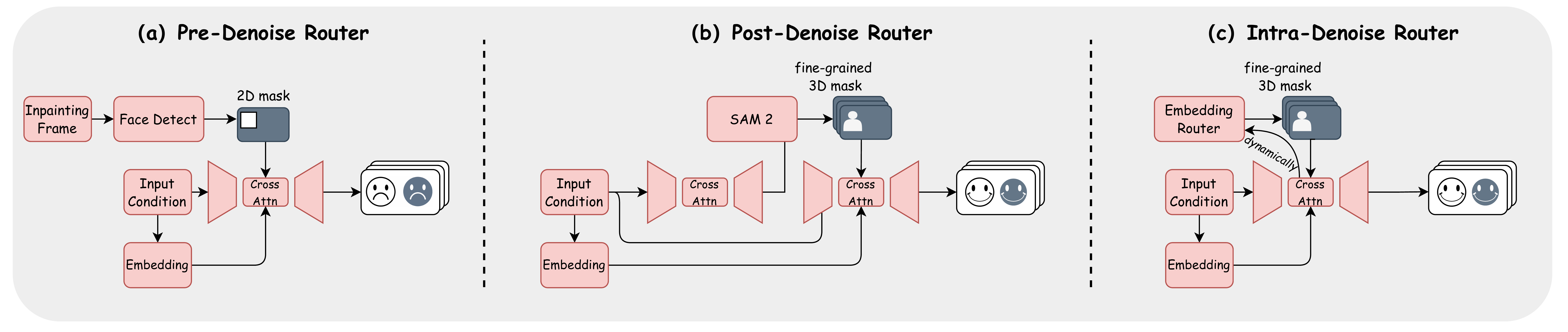} 
    \caption{Illustration of the three approaches to implement the \textbf{Embedding Router}: (1) \textbf{Pre-Denoise Router}, which uses a static visual mask extracted from the inpainting frame; (2) \textbf{Post-Denoise Router}, which detects face regions after the initial denoising process; and (3) \textbf{Intra-Denoise Router}, which dynamically generates visual masks during denoising using a learned module.}
    \label{fig2}
\end{figure*}

\section{Methodology}


\subsection{Task and Problem Formulation}
Given the following inputs:
A text prompt $\mathbf{x}$ describing the environment and the characters’ positions and actions; 
Multiple speech audios $\mathbf{a} = \{\mathbf{a}_1, \mathbf{a}_2, \dots, \mathbf{a}_n\}$, where $n$ is the number of characters, for driving each character's lip movements, facial expressions, and head poses; 
Multiple reference character images $\mathbf{I_r} = \{\mathbf{i_r}_1, \mathbf{i_r}_2, \dots, \mathbf{i_r}_n\}$, where $\mathbf{i_r}_i \in \mathbb{R}^{3 \times H \times W}$, for preserving characters' facial features; 
An optional inpainting frame $\mathbf{I_i} \in \mathbb{R}^{3 \times H \times W}$; 
An optional Audio-Character Matrix $\mathbf{A}^{\text{ac}} \in \mathbb{R}^{n \times n}$ indicating the correspondence between audios and characters; 
The goal is to generate a video sequence $\mathbf{V} \in \mathbb{R}^{T \times 3 \times H \times W}$ featuring multiple talking characters, where each character's lip movements and head poses are synchronized with their respective input speech audio, and their identities are accurately preserved based on the provided reference images.

\subsection{Framework}
\label{framework}
Our framework is primarily built upon CogVideoX \cite{yang2024cogvideox}, a multimodal diffusion transformer (MM-DiT) text-to-video foundational model. During the training process at timestep $t$, we first encode the video sequence $\mathbf{V}$ into a latent space $\mathbf{z_{0}} \in \mathbb{R}^{T' \times C' \times H' \times W'}$ using a pre-trained 3D VAE Encoder \cite{yang2024cogvideox}, where $C'$ is the number of channels, $T'$, $H'$, and $W'$ are the number of frames, height, and width of the latent space, respectively. 
Similarly, we encode the optional inpainting frame $\mathbf{I_i}$ and the concatenated reference images $\mathbf{I_{R}}$ (created by combining multiple reference character images $\mathbf{I_r}$ in the RGB space) into latent spaces. These latent representations are then padded along the time dimension to match the length of $\mathbf{z_{0}}$. Finally, we concatenate these three latent representations to form $\mathbf{z_{0}} \in \mathbb{R}^{T' \times 3C' \times H' \times W'}$.
Next, we divide the latent representation $\mathbf{z_{0}}$ into patches to obtain a sequence of 
visual tokens \( \mathbf{v} \in \mathbb{R}^{S \times d} \), where \( S \) is the sequence length and \( d \) is the inner dimension of the transformer block. The sequence length \( S \) is calculated as \( S = T' \cdot H' / \tau \cdot W' / \tau \), where \( \tau \) represents the patch size. 
These tokens are concatenated with the text embeddings and passed through $L$ layers of diffusion transformer blocks to predict the noise $ \epsilon_\theta $, while $ \epsilon $ is the noise added to $ v $, where $ \epsilon $ is sampled from a standard normal distribution (i.e., $ \epsilon \sim \mathcal{N}(0, 1) $).
Finally, we compute the diffusion loss to optimize the model:
\begin{equation}
    \mathcal{L}_{d} = \mathbb{E}_{v, t, \epsilon, x, I_r, a} \left[ \left\| \epsilon - \epsilon_\theta\left(v, t, \tau_\beta(x), \tau_\theta(I_r), \tau_\gamma(a)\right) \right\|_2^2 \right],
\end{equation}
where \( \tau_\beta(\cdot) \) denotes the text encoder, \( \tau_\theta(\cdot) \) denotes the face encoder, and \( \tau_\gamma(\cdot) \) denotes the audio encoder.

\noindent\textbf{Condition Injection}
As illustrated in Figure~\ref{fig1}, to effectively inject conditions into the denoising process, we utilize a T5 Encoder \cite{raffel2020exploring} to encode the text condition. For the audio condition, we first employ a pre-trained Wav2Vec \cite{baevski2020wav2vec} encoder to extract audio features, which are then projected to a higher-dimensional space. To address the temporal resolution misalignment between audio features and visual features, we apply a 2D convolution to down-sample the audio features, ensuring alignment with the temporal resolution of the visual features. The resulting audio representation is denoted as the audio embedding $\mathbf{e_a}$. 
To preserve character identities, we follow \cite{yuan_identity-preserving_2024}, utilizing a pre-trained CLIP model \cite{radford2021learning} to extract global features and a pre-trained InsightFace model \cite{deepinsight2024insightface} to extract local facial features. These features are then fused using a Q-Former \cite{li2023blip} to obtain refined features, which we denote as the character facial embedding $\mathbf{e_{c}}$.
Next, we encode both the character reference images and the inpainting frame through the 3D VAE encoder to obtain their latent representations. To better leverage their respective information content, we first apply noise to the facial regions of the inpainting frame and remove the background from the character reference images, both prior to encoding.
Finally, we apply mask-guided cross-attention to selectively integrate the embeddings into the visual tokens, as detailed in \ref{subsec:embedding_router}.
\subsection{Embedding Router}
\begin{figure}[t]
    \centering
    \includegraphics[width=0.5\textwidth]{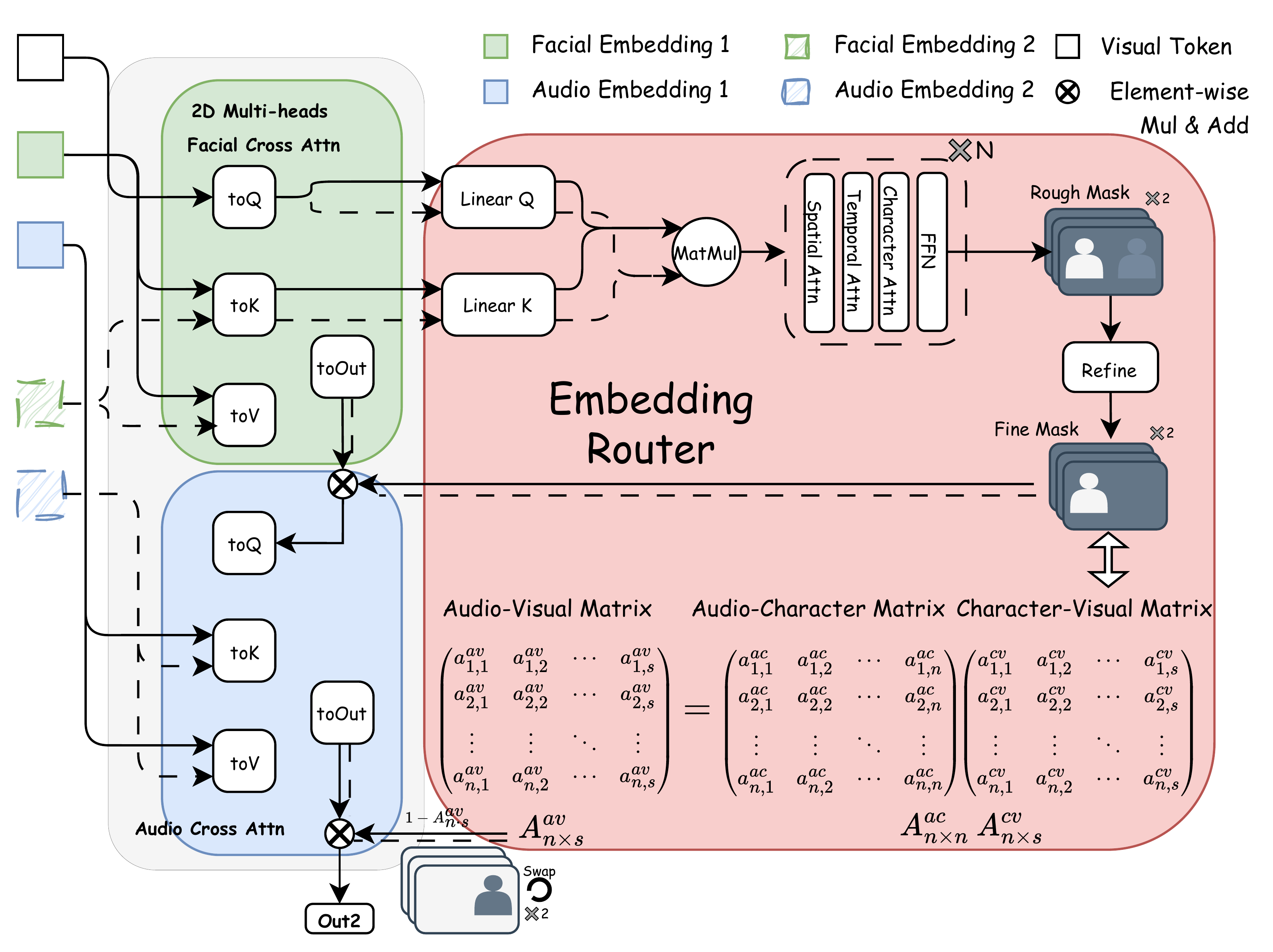} 
    \caption{Architecture of the proposed Intra-Denoise Router. Multi-character facial and audio embeddings (\textcolor[HTML]{82B366}{green} and \textcolor[HTML]{6C8EBF}{blue} square on the left) utilize rich cross-modal representations from pretrained cross-attention, followed by a lightweight transformer to capture global dependencies and refine the mask. The resulting fine mask and audio-visual matrix, computed by matrix multiplication, are used to selectively inject embeddings into the corresponding characters.}
    \label{fig3}
\end{figure}
\label{subsec:embedding_router}
To address the audio-to-character correspondence control in the same scene, we design an embedding router that binds speaker identity (``who'') with speech content (``what'') together.
We first present the overall pipeline of the embedding router, as illustrated in Figure \ref{fig2}, highlighting our proposed 3D-mask-based approaches for handling dynamic scenes and close character interactions. We then introduce weakly supervised loss functions grounded in geometric priors to guide accurate mask prediction. Finally, we describe the network architecture and implementation details of the complete embedding router pipeline.
\subsubsection{Overall Pipeline}
We believe utilizing a spatiotemporal mask $\mathbf{M} \in [0, 1]^{n \times T' \times H'/ \tau \times W'/ \tau}$ predicted by embedding router to guide the condition injection process is an effective approach for achieving precise audio-to-character correspondence control. This ensures that each character's facial and speech audio embeddings are accurately routed to their respective visual tokens within the DiT blocks.
We first attempt to obtain a 2D mask by detecting facial regions in the inpainting frame $\mathbf{I_{R}}$, which is a common practice in talkinghead generation~\cite{cuiHallo2LongDurationHighResolution2024}. This mask is then downsampled and temporally replicated to align with the dimensions of the visual tokens \( \mathbf{v} \). Then we utilize a mask-guided cross-attention module to inject the embeddings into the DiT blocks, as shown in Figure~\ref{fig2} (a).
We refer to this method as the ``Pre-Denoise Router'', since the mask is predicted prior to the denosing process. 
However, this approach fails to achieve precise audio-to-character binding because the static mask cannot adapt to dynamic scenes where character positions change across frames. Besides, the prior-shaped bounding box detection is not fine-grained enough to handle cases where two characters are close together, leading to inter-character confusion.

To address these shortcoming, we propose a 3D-mask-based embedding router and present two methods for its implementation, as shown in Figure~\ref{fig2} (b,c).
Method (b), termed as the Post-Denoise Router (coarse-generation then mask-prediction), employs a two-stage strategy: it first performs denoising without conditioning embeddings to generate a coarse, speech-free video; then predicts fine-grained 3D masks from the generated frames for subsequent refinement denoising.
Although this method achieves more precise audio-to-character binding, it incurs significant computational overhead due to the need for multiple forward passes during inference.
To further enhance efficiency, we introduce the Intra-Denoise Router (dynamic intra-generation prediction), which trains an embedding router module that dynamically predicts fine-grained 3D masks during the denoising process, eliminating multi-pass computation while maintaining accuracy, as shown in Figure \ref{fig2} (c).
\subsubsection{Loss Function for Intra-Denoise Router}
The Intra-Denoise Router, which predicts masks based on the correlation between high-dimensional facial embeddings and visual tokens, is trained using a combination of cross-entropy loss and weakly supervised losses that incorporate geometric priors.
Unlike \cite{fei_ingredients_2025}, we reformulate the router module's objective as a multi-class classification task by treating the background as an additional class ($n + 1$). This formulation mitigates the impact of noisy outputs in background regions during inference.
Given the visual tokens \( \mathbf{v} \) from transformer layer \( l \) and multiple character reference images \( \mathbf{I_r} \), the router module predicts a 3D visual mask \( \mathbf{M}\), indicating the character each visual token corresponds to. The router module is mainly optimized using a cross-entropy loss function:

\begin{equation}
\mathcal{L}_r = \sum_{l,i,t,h,w=1}^{L, \, n, \, T', \, H'/\tau, \, W'/\tau} -y_{l, i, t, h, w} \log(\hat{y}_{l, i, t, h, w}),
\end{equation}
where \( y_{l, i, t, h, w} \) represents the ground truth mask value for Character \( i \) at Layer \( l \), Time \( t \), Height \( h \), and Width \( w \), which is extracted using a pretrained SAM2 \cite{ravi2024sam2} model and downsampled to match the size of \( \mathbf{v} \). The predicted mask value from the router module is denoted as \( \hat{y}_{l, i, t, h, w} \).

Compared to \cite{fei_ingredients_2025}, which treats each visual token as an isolated entity and struggle to produce a smooth mask, we incorporate \textbf{geometric priors} which naturally constrain the shape of 3D mask to enhance mask smoothness and consistency, followed by three constraints: (1) Spatiotemporal Consistency, (2) Layer-wise Consistency, (3) Identity Exclusivity.
In light of the geometric priors, we further design some weak supervision losses to regularize the training of the router module:

\begin{equation}
\mathcal{L}_{\text{st}} = \sum_{l=1, \, i=1}^{L, \, n} \left\| \nabla_{\text{Spatiotemporal}} M_{l,i, t} \right\|_1,
\end{equation}
\begin{equation}
\mathcal{L}_{\text{layer}} = \sum_{i=1, \, t=1,\, h=1, \, w=1}^{n, \, T',\, H'/\tau,\,W'/\tau} \text{Var}\left( \{ M_{l,i,t,h,w} \}_{l=1}^{L} \right).
\end{equation}
Here, \(\nabla_{\text{Spatiotemporal}} M_{l,i, t}\) denotes the discrete spatiotemporal gradient of the mask M, encompassing variations along the horizontal, vertical, and temporal axes.
We observed that the loss term \(\mathcal{L}_r\) inherently enforces Identity Exclusivity, ensuring that only one character is predicted at any given temporal-spatial position.
The overall loss to optimize the router module is defined as \( \mathcal{L}_{\text{router}} \):

\begin{equation}
\mathcal{L}_{\text{router}} = \mathcal{L}_r + \lambda_{st} \mathcal{L}_{st} + \lambda_l \mathcal{L}_{\text{layer}},
\end{equation}
where \( \lambda_{st} \), and \( \lambda_l \) are the loss weights.

\subsubsection{Architecture Implementation}
As shown in Figure \ref{fig3}, Our router module architecture integrates both Geometric Priors and rich cross-modal representations from facial cross-attention, which is pre-trained in the first stage (Section \ref{subsec:training}). Details about network structure are provided in the Appendix.

During inference, we refine predicted masks by discarding low-confidence predictions and applying semi-supervised clustering for temporal-spatial consistency. Inspired by \cite{leonardis_emo_2025}, to allow more flexibility in character motion, we inflate the audio cross-attention masks by flipping their values and swapping along the character dimension.

After obtaining the mask M of Layer l, which can be viewed as the Character-Visual Matrix $\mathbf{A}^{\text{cv}} \in \mathbb{R}^{n \times s}$ (where $s = T' \cdot H'/\tau \cdot W'/\tau$ is the sequence length of visual tokens), we can derive the Audio-Visual Matrix $\mathbf{A}_{\text{av}}$ through:
\begin{equation}
\mathbf{A}^{\text{av}} = \mathbf{A}^{\text{ac}} \cdot \mathbf{A}^{\text{cv}},
\end{equation}
where $\mathbf{A}_{\text{ac}} \in \mathbb{R}^{n \times n}$ is the Audio-Character Matrix indicating the correspondence between audios and characters. 
This matrix can either be provided as an input condition or predicted by our proposed audio router module. In this paper, we implement the audio router by leveraging a pretrained speech-face alignment model \cite{wen_seeking_2021} to compute relevance scores between audio streams and characters, followed by a voting mechanism to derive the final Audio-Character Matrix. Finally, we apply mask-guided cross-attention in two sequential steps to integrate both the facial embeddings and the motion-related speech information into the visual tokens by:
\begin{equation}
    \mathbf{v'} = \mathbf{v} + \mathbf{A}^{\text{cv}} \cdot \texttt{FaceCrossAttn}(\mathbf{e_{c}}, \mathbf{v}),
\end{equation}
\begin{equation}
    \mathbf{v''} = \mathbf{v} + \mathbf{A}^{\text{av}} \cdot \texttt{AudioCrossAttn}(\mathbf{e_{a}}, \mathbf{v'}),
\end{equation}
where \(\mathbf{e_{c}}\) and \(\mathbf{e_{a}}\) denote the character facial embedding and audio embedding, respectively, as described in Section~\ref{framework}.

\subsection{Multi-Stage Training}
\label{subsec:training}
We divide the training process into three consecutive stages to progressively enhance the model's capabilities, from identity preservation to audio-driven motion control, and finally to multi-character audio-driven animation.

\textbf{Stage 1:} Excluding speech audio conditions, we focus on identity preservation and the ability to generate from an inpainting frame. The full parameters of the transformer and face embedding extractor are unfrozen. We introduce a 50\% probability of randomly dropping the inpainting frame, forcing the model to generate robust outputs in both conditional and unconditional scenarios. Following \cite{yuan_identity-preserving_2024}, a Dynamic Mask Loss with a 50\% application rate is adopted.

\textbf{Stage 2:} Audio conditions are added, and Low-Rank Adaptation (LORA) \cite{hu_lora_2021} is incorporated into the video DiT architecture as an additional trainable module. All parameters are fixed except for the audio encoder, audio cross-attention, face encoder, face cross-attention, and the DiT's LORA. Random drops of different condition combinations are applied to strengthen classifier-free guidance generation \cite{ho2022classifier}.

\textbf{Stage 3:} The embedding router module is introduced and trained jointly with the entire denoising process. All parameters are fixed except for the embedding router module, audio cross-attention, face cross-attention, and the DiT's LORA. 
We adopt a two-task hybrid training paradigm by dividing the training process into single-character and multi-character animation tasks. These tasks use distinct training data while sharing the same network parameters. For single-character scenarios, we replicate the input conditions \(n\) times to match the model's expected multi-character input format.
We also employ a teacher-forcing training strategy to stabilize the training process, as detailed in the appendix.
\begin{figure*}[t]
    \centering
    \includegraphics[width=1.0\textwidth]{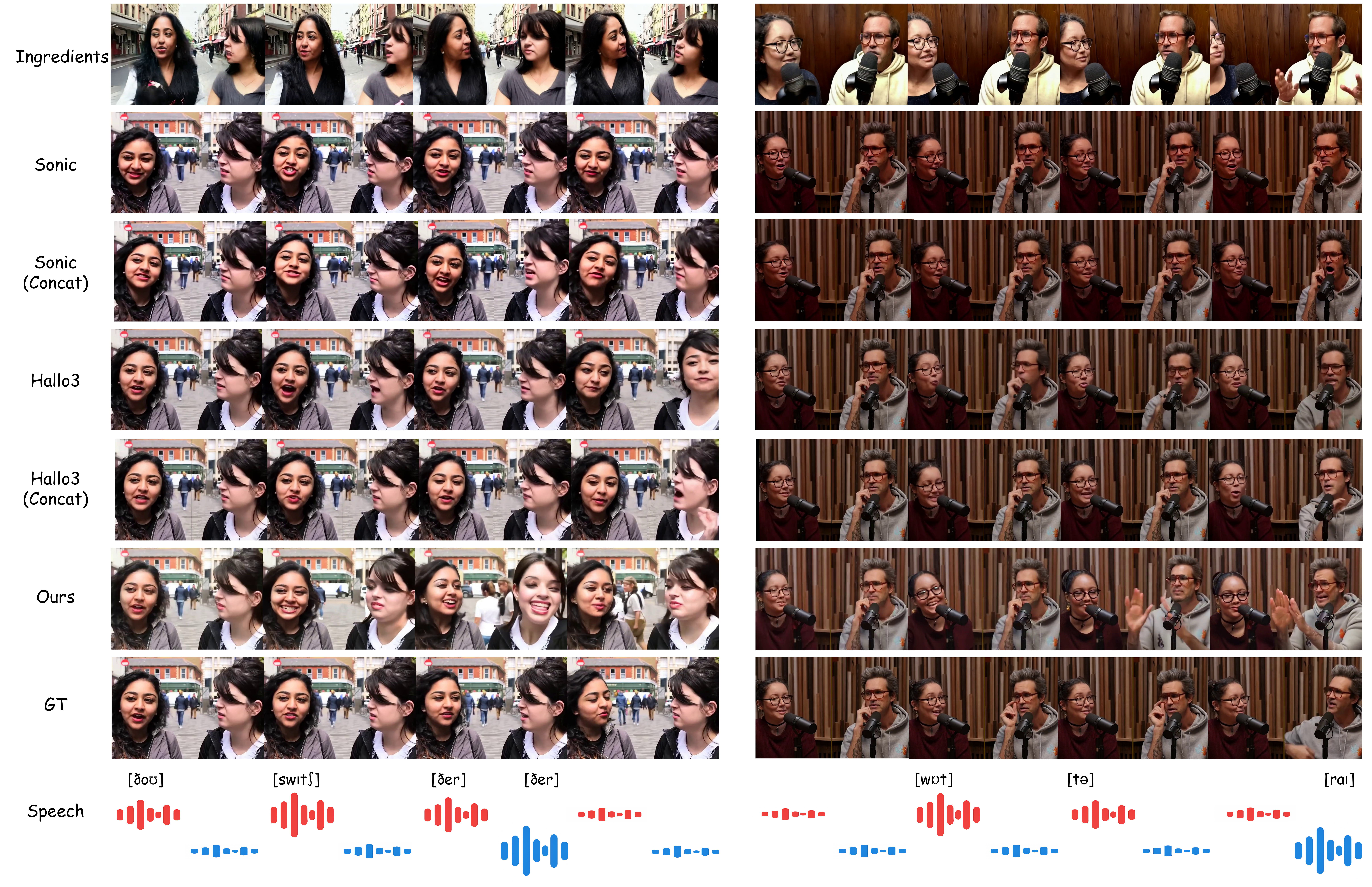} 
    \caption{
    Qualitative comparison with other competing methods on MTCC test set.
    }
    \label{fig5}
\end{figure*}
\section{Dataset}
To address the challenges of the lack of datasets for multi-talking-character videos in the same scene, we introduce a novel dataset, \textbf{MTCC} (Multi-Talking-Characters-Conversations), specifically designed for multi-talking-character video generation. 
This dataset contains over 200 hours of raw video footage, curating from various sources, including talk shows, interviews, news, online lessons, and TV series, to ensure a diverse range of characters, scenes, and interactions.
\subsection{Data Cleaning}
We apply a series of automated data cleaning steps to ensure dataset quality—a stage we deem crucial for training. First, we filter raw videos by resolution, duration, and FPS. Then we employ a face detection model \cite{yolov8_ultralytics} to extract bounding box and isolate segments containing exactly two characters.
Next, we calculate the bounding-box-to-inter-character-distance ratio to exclude segments where the characters' bodies are too distant. Finally, lip-sync quality filtering is performed.
Directly applying Sync-C~\cite{chung2017out} to segments generates noisy outputs due to overlapping speech from both characters. Consequently, segments with simultaneous dialogue are mistakenly filtered out. To overcome this, we utilize the AV\_MossFormer2\_TSE\_16K model \cite{clearvoice_2025} to separate speech tracks and remove irrelevant noise. Sync-C is computed individually for each character's isolated audio and cropped facial video, the average of these two scores serves as our filtering criterion.
\subsection{Processing Pipeline}
To enable multi-character video training, we construct a comprehensive preprocessing pipeline following data cleaning. This pipeline includes segment extraction, speech separation, audio embedding extraction, and caption generation. Specifically, we use SAM2~\cite{ravi2024sam2} to obtain per-character segmentation masks, which are downsampled to the latent space and used as ground truth for the routing masks. AV\_MossFormer2\_TSE\_16K is applied to separate speech tracks from overlapping videos, establishing the Audio-Character Matrix $\mathbf{A}_{\text{ac}}$ that encodes the correspondence between audio and characters. Audio embeddings are extracted from each isolated speech segment using Wav2Vec~\cite{baevski2020wav2vec}. Finally, QWEN2-VL~\cite{bai2023qwen} is employed to generate descriptive captions for all segments.

\section{Experiments}
\subsection{Bind-Your-Avatar-Benchmark}
To the best of our knowledge, no publicly available benchmark has been specifically designed for multi-talking-character scenarios. To comprehensively assess the model's generalizability in animating arbitrary characters in diverse real-world scenarios, we introduce Bind-Your-Avatar-Benchmark, a novel benchmark. We've meticulously collected 40 pairs of portraits from various domains, featuring an eclectic mix of styles, such as Tiefling-like fantasy creatures and sci-fi-inspired humanoid robots. These portraits showcase a wide age range, including adults, young people, teenagers, and infants, set against backgrounds that span from realistic settings to fantastical and science-fiction landscapes.
Each portrait pair is accompanied by corresponding dual-stream speech audio clips, generated by the text-to-speech model \cite{du2024cosyvoice2}, as well as textual prompts. These prompts describe different appearances (e.g., wearing sunglasses, in a black suit), actions (e.g., reach out, head shaking), and environments (e.g., The view out the window is changing). The benchmark comprises two subsets: 20 pairs come with corresponding inpainting frames, while the remaining 20 pairs do not, offering versatility for different research needs and expected to significantly benefit the development of the community.
\subsection{Experimental Setup}
\begin{figure*}[t]
    \centering
    \includegraphics[width=1\textwidth]{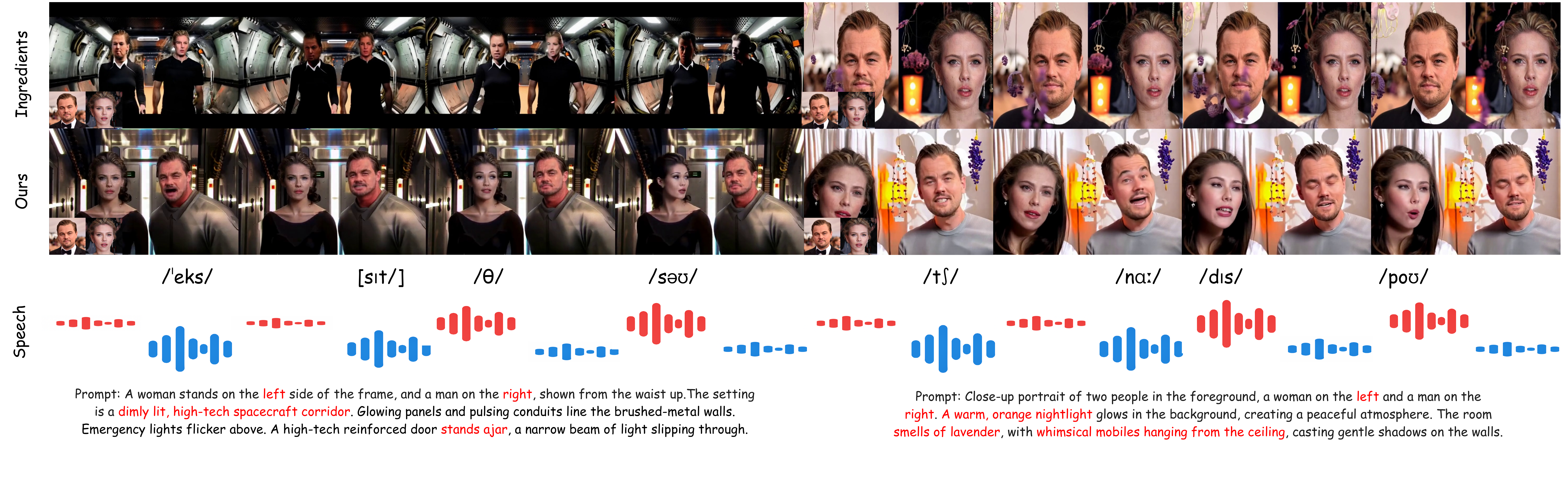} 
    \caption{
    Qualitative comparison with Ingredients~\cite{fei_ingredients_2025} on the Bind-Your-Avatar-Benchmark (without inpainting frame), which shows that our method have strong ability on both content detail generation and identity preservation.
    }
    \label{fig7}
\end{figure*}
\subsubsection{Implementation Details}
For simplicity, we set the number of characters $n$ to 2 in all experiments. Across all training phases, the three stages consist of 10,000, 40,000, and 10,000 steps, respectively. The total batch size is set to 16, and the learning rate is $1 \times 10^{-5}$. To enhance classifier-free guidance, the reference image, inpainting frame, and audio are each randomly dropped with a probability of 0.05 during training.
The loss weights for $\mathcal{L}_r$, $\mathcal{L}_{st}$, and $\mathcal{L}_{\text{layer}}$ are set to 1, 0.001, and 8, respectively.
During inference, we use 50 sampling steps and set the classifier-free guidance (CFG) scale for both audio and text to 7.
Our training sets include: (1) single-character datasets: Hallo3 \cite{cuiHallo3HighlyDynamic2025}, HDTF \cite{zhang2021flow}, VFHQ \cite{xie2022vfhq}; and (2) multi-character datasets: VICO \cite{zhou2022responsive}, Friends-MMC \cite{wang2025friends}, and our dataset. VFHQ is curated using URL labels due to its silent nature. For VICO, we concatenate per-character videos to form unified multi-character scenes. Friends-MMC is cleaned and preprocessed as with our dataset.

\subsubsection{Evaluation Metrics}
We evaluate all methods based on three aspects: identity preservation, audio-visual synchronization, and visual quality.
\textbf{Identity Preservation:} Following \cite{yuan_identity-preserving_2024,fei_ingredients_2025}, we compute pairwise face similarities (Face Sim) between reference and generated faces using FaceNet \cite{schroff2015facenet}.
\textbf{Audio-Visual Synchronization:} For each character, we use a face detector \cite{yolov8_ultralytics} to crop faces and apply AV\_MossFormer2\_TSE\_16K \cite{clearvoice_2025} for speech separation. We then calculate Sync C and Sync D scores, which are metrics that quantify audio-visual sync quality, by averaging across all characters.
\textbf{Visual Quality:} FID~\cite{heusel2017gans} (on face regions using InceptionV3~\cite{szegedy2016rethinking}) and FVD~\cite{unterthiner2019fvd} are used to evaluate generation quality.
\subsubsection{Test Sets}
For the multi-talking-characters video generation task, we employ two test sets. The first comprises 20 clips derived from our proposed MTCC datasets. The second test set utilizes the 20 pairs from the Bind-Your-Avatar-Benchmark that include inpainting frames.
For the multi-talking-characters video generation (without inpainting frame) task, we use the 20 pairs from the Bind-Your-Avatar-Benchmark that lack inpainting frames. These carefully selected test sets, in conjunction with the benchmark, allow for a thorough and nuanced evaluation of our method across different video generation tasks.
\begin{figure}[t]
    \centering
    \includegraphics[width=0.47\textwidth]{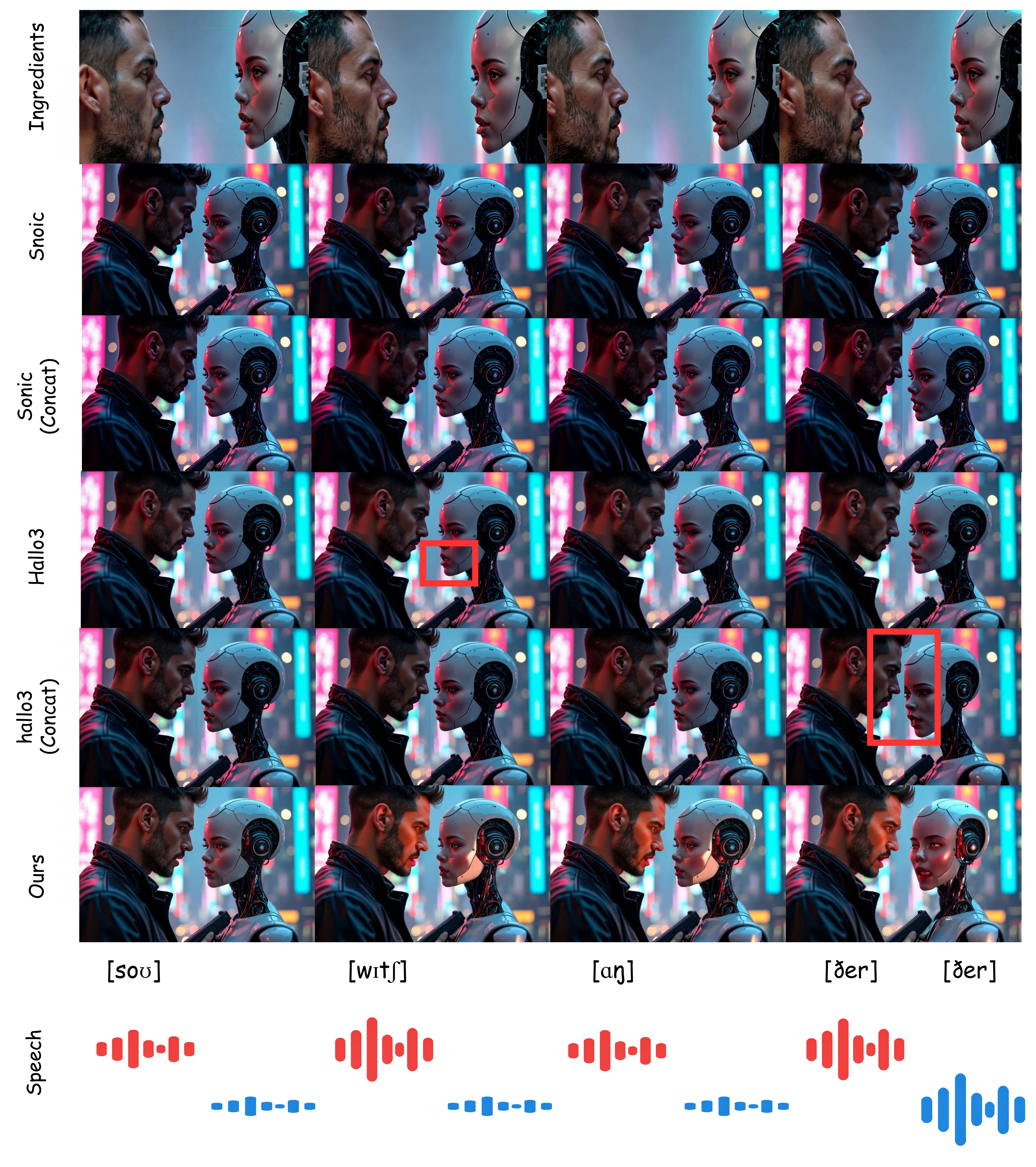} 
    \caption{Qualitative comparison with other competing methods on the Bind-Your-Avatar-Benchmark (with inpainting frame).}
    \label{fig6}
\end{figure}
\subsection{Comparison with State-of-the-Art}
\subsubsection{Quantitative Evaluation}
To verify the effectiveness of our method, we compare it with several state-of-the-art character animation approaches. As no existing method specifically targets multi-talking-character task, to the best of our knowledge, we select representative baselines from related areas, including traditional talking-head generation methods (Sonic~\cite{jiSonicShiftingFocus2024} and Hallo3~\cite{cuiHallo3HighlyDynamic2025}) and a multi-concept human video generation method (Ingredients~\cite{fei_ingredients_2025}).

Since Hallo3 and Sonic are not designed to handle multi-stream speech audio, we adopt two practical adaptations for implementation: (1) directly using the overlapping speech as input, or (2) providing two separate clean speech streams along with the left and right regions of the inpainting frame as inputs to generate two independent videos, which are then concatenated. The latter approach corresponds to the (Concat) variant illustrated in Figure~\ref{fig5}. Additionally, as Hallo3 and Sonic depend on inpainting frames, they are not included in the multi-talking-character (without inpainting frame) comparison. As for Ingredients, since it does not support audio or inpainting frame inputs, we omit these modalities and append the corresponding driving audio to the generated video for Sync-C and Sync-D evaluation.
\begin{table}[t]
    \centering
    \caption{Quantitative comparisons in multi-talking-characters video generation task on MTCC test set. The best results are in \textbf{Bold}.}
    \label{table2}
    \fontsize{6}{6.5}\selectfont
    \setlength{\tabcolsep}{3pt}
    \renewcommand{\arraystretch}{1.4}
    \begin{tabular}{lC{1.5cm}C{1cm}C{1cm}C{0.9cm}C{0.9cm}}
        \hline\noalign{\hrule height 0.6pt}
        \textbf{Method}
        & \textbf{Face Sim~$\uparrow$(\%)}
        & \textbf{Sync-C~$\uparrow$}
        & \textbf{Sync-D~$\downarrow$}
        & \textbf{FID~$\downarrow$}
        & \textbf{FVD~$\downarrow$} \\
        \hline 
        Ingredients& 48.21&	1.02	&12.22&	122.16	&922.31 \\ 
        Sonic & \textbf{76.92}&	2.25	&9.76	&29.98	&312.12\\ 
        Sonic (Concat) & 75.89&	4.52&	8.89&	36.78&	355.12 \\ 
        Hallo3 & 70.14	&2.15	&10.23&	31.26&	332.68 \\ 
        Hallo3 (Concat)&70.87&	4.12&	8.98	&41.25&	422.63\\ 
        \textbf{Ours} &71.26	& \textbf{4.67}&	\textbf{8.58}&	\textbf{29.55}&	\textbf{308.67} \\ 
        \hline\noalign{\hrule height 1.2pt}
    \end{tabular}
\end{table}
\begin{table}[t]
    \centering
    \caption{Quantitative comparisons in multi-talking-characters video generation on Bind-Your-Avatar-Benchmark. The best results are in \textbf{Bold} and the second best results are \underline{underlined}.}
    \label{table3}
    \fontsize{7.5}{6.5}\selectfont
    \setlength{\tabcolsep}{3pt}
    \renewcommand{\arraystretch}{1.4}
    \begin{tabular}{lC{1.77cm}C{1.77cm}C{1.77cm}}
        \hline\noalign{\hrule height 0.6pt}
        \textbf{Method}
        & \textbf{Face Sim~$\uparrow$(\%)}
        & \textbf{Sync-C~$\uparrow$}
        & \textbf{Sync-D~$\downarrow$} \\
        \hline
        Ingredients & 51.22	&0.95	&12.97 \\ 
        Sonic &\textbf{71.50}	&2.32	&9.82  \\ 
        Sonic (Concat) & 64.78	&\textbf{4.39}&	8.89  \\ 
        Hallo3 & 64.67	&2.12	&10.55  \\ 
        Hallo3 (Concat) & 61.29	&3.79&	8.72\\ 
        \textbf{Ours} & \underline{65.78}	&\underline{4.09}	&\textbf{8.68} \\ 
        \hline\noalign{\hrule height 0.6pt}
    \end{tabular}
\end{table}
Quantitative comparisons on both the multi-talking-character task and the multi-talking-character (without inpainting frame) task are presented in Table~\ref{table2},~\ref{table3}, and Table~\ref{table4}, respectively. Our method outperforms most baselines across the majority of metrics, demonstrating superior audio-visual synchronization and identity preservation.
\subsubsection{Qualitative Evaluation}
To demonstrate the visual effectiveness of our method, we compare and visualize the results alongside several competitive approaches, as shown in Figure~\ref{fig5} and Figure~\ref{fig6} for the multi-talking-character task, and in Figure~\ref{fig7} for the multi-talking-character (without inpainting frame) task. The red highlighted box in Figure~\ref{fig6} indicates a failure case of the competing methods in responding to two-stream speech input and generating a unified background, which our method successfully addresses. Moreover, our method is capable of producing dynamic and vivid backgrounds—with natural lighting variations and scene dynamics such as moving pedestrians in the distance and subtle light changes on the speaker’s face—whereas other state-of-the-art methods tend to produce static and less expressive scenes. This advantage stems from the diverse, in-the-wild multi-speaker scenarios provided by our proposed MTCC dataset.
As shown in Figure~\ref{fig7}, our method not only effectively captures environment details and spatial relationships between characters as described in the prompt, but also maintains strong identity preservation and precise correspondence between each character and their respective speech stream.
\begin{table}[t]
    \centering
    \caption{Quantitative comparisons in multi-talking-character (without inpainting frame) task on Bind-Your-Avatar-Benchmark.}
    \label{table4}
    \fontsize{7.5}{6.5}\selectfont
    \setlength{\tabcolsep}{3pt}
    \renewcommand{\arraystretch}{1.4}
    \begin{tabular}{lC{1.91cm}C{1.91cm}C{1.91cm}}
        \hline 
        \textbf{Method}
        & \textbf{Face Sim~$\uparrow$(\%)}
        & \textbf{Sync-C~$\uparrow$}
        & \textbf{Sync-D~$\downarrow$} \\
        \hline
        Ingredients & 48.2 & 1.02 & 12.22  \\ 
        \textbf{Ours} & \textbf{68.9} & \textbf{4.52} & \textbf{7.89}  \\ 
        \hline\noalign{\hrule height 0.6pt}
    \end{tabular}
\end{table}

\subsection{Ablation Studies}
\subsubsection{Bounding Box vs Fine-grained Mask}
\begin{figure}[t]
    \centering
    \includegraphics[width=0.49\textwidth]{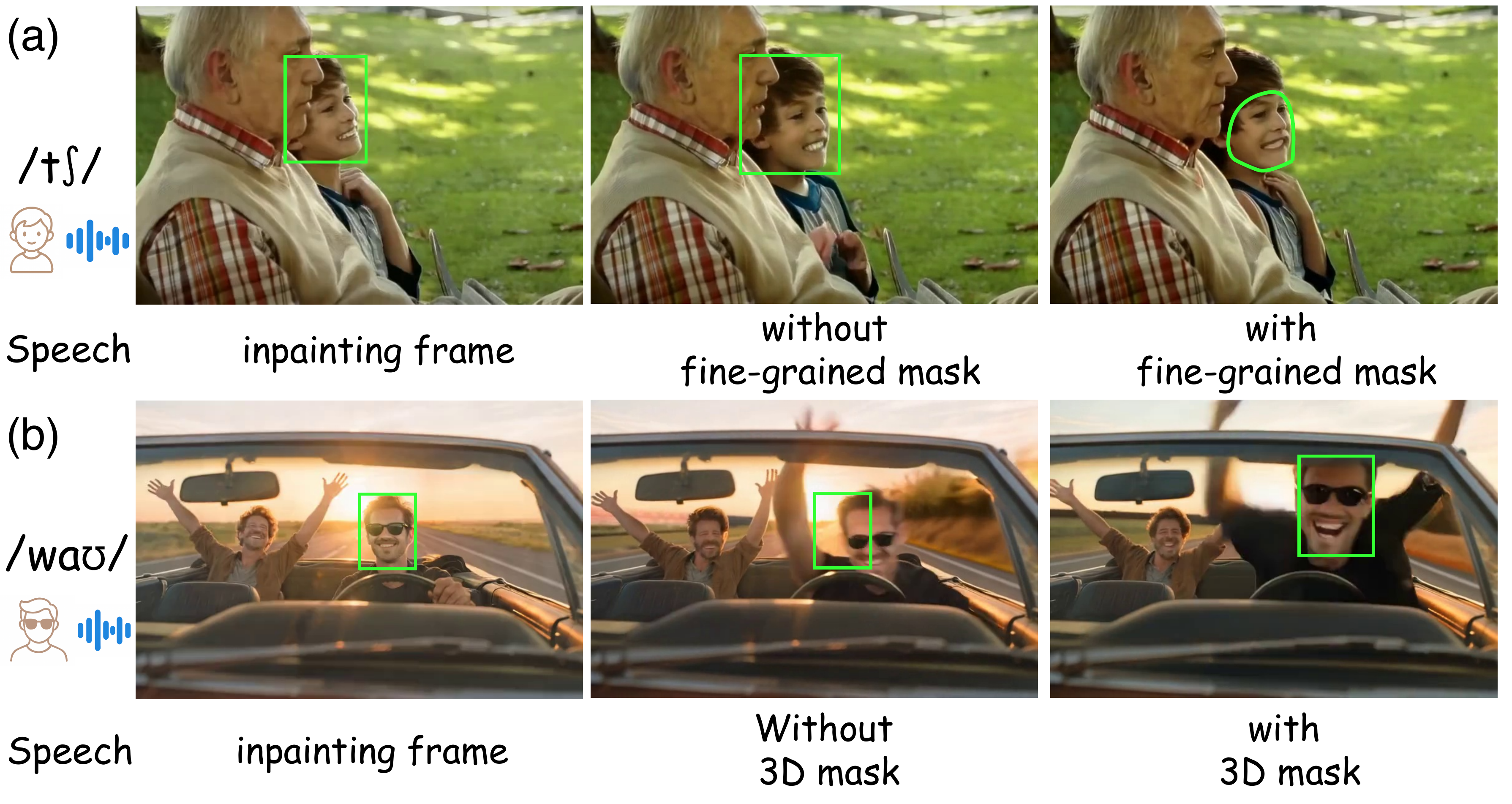} 
    \caption{Ablation on (a) bounding box vs fine-grained mask, and (b) 2D mask vs 3D mask.}
    \label{fig8}
\end{figure}

To investigate whether our proposed fine-grained mask yields more precise results, we compare its performance with traditional bounding box masks in close-interaction scenes, a common practice in talking-head methods~\cite{cuiHallo2LongDurationHighResolution2024}. Figure~\ref{fig8}(a) demonstrates cases where our fine-grained mask outperforms bounding boxes. In close interactions, overlapping bounding boxes fail to correctly bind audio to characters, while our dense masks accurately separate facial regions for precise correspondence.
\subsubsection{2D Mask vs 3D Mask}  
To evaluate the role of spatiotemporal mask adaptability in dynamic scenes, we contrast static 2D masks (extracted from the inpainting frame and fixed across frames) with our dynamic 3D masks. As shown in Figure \ref{fig8}(b), in a driving scenario with character motion, the 2D mask—anchored to the initial frame—fails to track the driver’s posture changes. This misaligns the mask with the actual face in subsequent frames, causing audio-character binding errors.
In contrast, our 3D mask dynamically updates frame-by-frame, aligning with the driver’s changing position to maintain precise audio-visual synchronization. This highlights that 3D masks are critical for dynamic scene modeling.
\section{Conclusion}
In this paper, we present Bind-Your-Avatar, an MM-DiT-based framework for multi-talking-character video generation in a shared scene. To address the challenge of audio-to-character correspondence, we propose a fine-grained embedding router that binds “who” and “speak what” together. We introduce two 3D-mask router implementations with geometry-based losses and mask refinement. To support this task, we construct the first dedicated dataset and release an open-source data processing pipeline. Extensive experiments on MTCC and our Bind-Your-Avatar-Benchmark demonstrate the effectiveness of our approach.
\subsubsection{Limitations}
Our model lacks explicit human motion modeling and is trained on limited audio-driven video data, which seldom include multi-person interactions. This results in suboptimal hand gestures and less realistic character interactions. Additionally, the full-attention architecture of MM-DiT leads to high inference costs, limiting real-time applicability. Future work will focus on improving motion fidelity and reducing computational overhead for practical deployment.
\appendix
\section{Intra-Denoise Router Architecture}
The network architecture of Intra-Denoise Router, grounded in the Geometric Priors described in Section~\ref{subsec:embedding_router}, leverages the rich cross-modal representations captured by pretrained facial cross-attention. Specifically, as shown in Figure \ref{fig3}, we utilize the query and key outputs from a pretrained multi-head face cross-attention module as inputs to the router module. Both query and key undergo layer-wise linear transformations followed by reshaping operations that preserve the head dimension. The attention weights, computed via matrix multiplication between transformed queries and keys, encode the correspondence between facial features and visual tokens. These weights are enhanced with 3D RoPE positional encoding~\cite{su2024roformer} before processing through multiple spatio-temporal attention layers that model interdependencies among visual tokens. A final linear layer with softmax activation generates the predicted mask output. This design holistically captures both isolated feature-token correspondences and spatio-temporal token dependencies, which are essential for producing precise, smooth masks.
\section{Teacher-Forcing Training Strategy}
We observed that training the router module jointly with the entire denoising process often leads to a trivial solution, where visual tokens become easy to classify but lack meaningful visual representation. Meanwhile, the router's training depends on the model being sufficiently adapted to the data distribution to perform high-quality denoising. Without accurate router predictions, the diffusion model progressively loses its ability to effectively integrate and utilize the provided conditions.

Thus,we employ a teacher-forcing training strategy, where we force the input of mask-guided cross attention in the transformer diffusion model to be the ground truth mask value, and detach the computational graph of the router module from the denoising process.
To enhance noise robustness, we apply a dropout operation and add Gaussian noise to the forced mask values, effectively serving as a form of data augmentation.
{
    \small
    \bibliographystyle{ieeenat_fullname}
    \bibliography{main}
}


\end{document}